\documentclass{article}
\usepackage{neurodata}

\usepackage{enumitem}
\usepackage{subcaption}
\usepackage{lipsum}
\usepackage{comment}
\usepackage{comment}
\usepackage{algpseudocode}
\usepackage{wrapfig}
\usepackage{dsfont}
\usepackage{bbm}

\title{Temporal Chunking Enhances Recognition of Implicit Sequential Patterns}

\author[1]{Jayanta Dey}
\affil[1]{University of Texas at San Antonio}

\author[1]{	
Nicholas Soures}

\author[1]{	
Miranda Gonzales}

\author[1]{	
Itamar Lerner}
\affil[2]{University of Rochester}

\author[2]{	
Christopher Kanan}

\author[1]{	
Dhireesha Kudithipudi\thanks{dhireesha.kudithipudi@utsa.edu}}

\begin{document}
\maketitle

\begin{abstract}
In this pilot study, we propose a neuro-inspired approach that compresses temporal sequences into context-tagged chunks, where each tag represents a recurring structural unit or ``community'' in the sequence. These tags are generated during an offline sleep phase and serve as compact references to past experience, allowing the learner to incorporate information beyond its immediate input range. We evaluate this idea in a controlled synthetic environment designed to reveal the limitations of traditional neural network based sequence learners, such as recurrent neural networks (RNNs), when facing temporal patterns on multiple timescales. Our results, while preliminary, suggest that temporal chunking can significantly enhance learning efficiency under resource constrained settings. A small-scale human pilot study using a Serial Reaction Time Task further motivates the idea of structural abstraction. Although limited to synthetic tasks, this work serves as an early proof-of-concept, with initial evidence that learned context tags can transfer across related tasks—offering potential for future applications in transfer learning.

\end{abstract}

\tableofcontents

\section{Introduction}

Recurrent neural networks (RNNs) are widely used for modeling sequential data, but their ability to capture long-range temporal structure is often constrained by practical limitations. While RNNs have a theoretically unbounded context window—since information can, in principle, persist within the hidden state \citep{hochreiter1997long}—this memory is lossy. Over time, signals tend to dissipate or become entangled \citep{bengio1994learning}, making it difficult to retain information from distant past events. Whether due to a fixed window or lossy dynamics, RNNs struggle with long-range dependencies, especially when events are separated by large or irregular time gaps. This issue is exacerbated when training with truncated backpropagation through time (BPTT) or short input windows, limiting the model's ability to utilize earlier information. As a result, RNNs often fail to capture the temporal regularities that span more steps than the model can access directly.

Ideally, we would evaluate such mechanisms in learning environments involving long-range dependencies. However, to study these challenges in a controlled and interpretable setting, we design a synthetic sequence modeling environment where the length of the temporal context required to infer the correct prediction is precisely specified. This setup allows us to directly test how well a model with constrained resources can recover implicit temporal rules. When the input window or the BPTT window is too short, standard RNNs are unable to resolve the necessary dynamics and perform poorly.

Intriguingly, the above constraints resonate with long-standing insights from neuroscience and psychology, where human cognition employs multiple strategies to cope with limited working memory. Notably, the well-studied phenomenon of chunking allows humans to group separate items or events into meaningful units, effectively increasing the capacity of short-term recall \citep{Miller1956, Norris2021}. Furthermore, evidence suggests that sleep-based memory consolidation might contribute to the formation of those chunks. Specifically, it is known that sequences of stored wake experiences are reactivated during sleep at a compressed timescale \citep{Mehta2007, Rasch2013}. According to the temporal scaffolding model of memory consolidation \citep{Lerner2017, lerner2019sleep}, such accelerated replay of past events during offline periods fosters the formation of long-range associations not easily captured during normal wake states alone. From a machine learning perspective, these mechanisms hint at a path to overcoming architectural context window limits: by chunking temporally extended data into higher-order representations, models can link distant events without explicitly expanding the underlying backpropagation through time (see Appendix~\ref{app:rnn} for details) or input window.

To explore this idea, we propose a pilot study of an offline temporal chunking mechanism that compresses temporal patterns into context-tagged chunks. Each context tag corresponds to the onset of a distinct temporal community (i.e., a group of states in the environment where the states evolves following a common temporal rule), and is learned during an offline phase by analyzing previously encountered temporal sequence. These tags are then reused during subsequent learning, effectively summarizing recurring structure and allowing the RNN to access relevant past information even when constrained to short input windows.

As a motivating parallel, we include a small-scale human pilot study using a Serial Reaction Time Task (SRTT) designed to mirror the synthetic setup. Although preliminary, the results suggest that participants may also form abstract representations of temporal structure after limited exposure. Additionally, we show initial evidence that context tags learned in one version of the task may remain useful after changes in the input dynamics, indicating potential for transfer learning, although this remains a direction for future exploration.

Overall, this work presents a proof-of-concept for context-based compression in time-series modeling. By isolating the core challenge of limited memory and using a synthetic environment to probe model behavior, we highlight the potential of temporal chunking-based context identification as a lightweight mechanism for learning implicit sequential pattern under resource constraints.

In summary, our main contributions are: 
\begin{enumerate}
    \item Motivated by the neuro-inspired temporal scaffolding hypothesis, we propose a temporal chunking-based framework that allows RNNs to detect temporal regularities extending well beyond their input window.
    \item We demonstrate that a three-stage learning model (reflecting pre-sleep wake, sleep, and post-sleep wake modes) can efficiently capture complex temporal patterns using far fewer BPTT steps while improving temporal learning performance.
    \item In a small-scale pilot human study using SRTT, we observe that participants show recognition of community structures after a brief training session. Although limited in scope, the results suggest that some degree of structural abstraction or ``chunking'' may emerge during wakeful learning alone, providing preliminary motivation for our proposed mechanism. A more controlled and extensive human study is left for future work.
\end{enumerate}

\section{Technical Background}

\subsection{Problem Setting}
\label{sec:problem}
Let $\{X_1, X_2, \cdots, X_n \}$ be a stochastic process or a sequence of random variables where each variable takes a value from some finite set $\mathcal{S} = \{s_1, s_2, \cdots, s_K\} \subset \mathbb{R}^d$. The variable $X_n$ represents the state of the process at time $n$, governed by a set of underlying state transition probability laws $\mathcal{P} = \{\mathbf{p}_1, \mathbf{p}_2, \cdots, \mathbf{p}_n\}$. Each $\mathbf{p}_n \in \mathcal{P}$ defines the probability of transitioning to any state $s_i \in \mathcal{S}, \forall i=1,\cdots,K$ given all previous states up to time $(n-1)$, that is, 

\begin{equation}
    \mathbf{p}_n = [p_n(s_i)]_{i=1}^K = [P(X_n=s_i|X_1,\cdots,X_{n-1})]_{i=1}^K.
\end{equation}

Given a sequence of states evolving according to an unknown transition rule $\mathcal{P}$, a sequential learner $f: S^T \rightarrow [0, 1]^K$, having access to an input window of past $T$ states, estimates $\mathbf{p}_n$ :

\begin{equation}
    \hat{\mathbf{p}}_n = [\hat{p}_n(s_i)]_{i=1}^K =  f(x_{n-T}, \cdots, x_{n-1} ),
\end{equation}

where $x_n$ is the value of the random variable $X_n$ at time $n$. The state at time $n$ is estimated as the $\argmax$ of $\hat{\mathbf{p}}_n$:

\begin{equation}
    \hat{x}_n = \argmax_{s_i \in \mathcal{S}} \hat{p}_n(s_i),  \forall i=1,\cdots,K.
\end{equation}
If the transition probability between states, $\mathbf{p}_n$, depends on more past states than those captured within the input window, then the learner must integrate an internal mechanism to retain the memory of the previous states. The estimation accuracy of the next state depends on how effectively the model retains and utilizes past information. Below we define`chunking' where we group learning rules based on their similarity to each other.


\textbf{Temporal Chunking:}
Let $\mathcal{P} = \{p_1, p_2, \dots, p_n\}$ denote a temporal sequence governed by an underlying transition structure, and let a \emph{community} be defined as a recurring substructure within the sequence that shares common temporal dynamics. We define \emph{chunking} as the process of decomposing $\mathcal{P}$ into a sequence of contiguous subsegments $\mathcal{P}_1, \mathcal{P}_2, \dots, \mathcal{P}_m$, where each $\mathcal{P}_i$ corresponds to a distinct community segment.

Each chunk $\mathcal{P}_i$ is assigned a \emph{context tag} $c_i$, which acts as a compressed latent representation or index of that community. The complete sequence can now be represented as a sequence of context tagged chunks:
\[
\mathcal{P} = \bigoplus_{i=1}^{m} (\mathcal{P}_{ c_i}),
\]
where $\oplus$ denotes ordered concatenation. The context tags $\{c_1, c_2, \dots, c_m\}$ abstract away the fine-grained transitions within each chunk, enabling the model to reason over a \emph{higher-level sequence of communities}.

Importantly, the chunking mechanism not only facilitates recall of local temporal structure within each chunk, but also allows the model to \emph{learn patterns over the sequence of context tags themselves}. In this way, chunking transforms the original sequence into a hierarchical representation, supporting efficient compression, memory utilization, and sequential learning under resource constraints.


 Chunking can be considered similar to the `associative memory' mechanism in biological agents where a certain cue or stimulus can invoke a chunk of memory or experience in the agent \citep{anderson2014human}. In addition, chunking can be related to mnemonics that are known to help memorization \citep{putnam2015mnemonics}, for example, `ROYGBIV' which is a popular mnemonic for remembering the rainbow colors. Each letter in the above mnemonics can be considered a context tag that invokes a certain color in our memory. Our proposed temporal chunking strategy, unlike the examples mentioned earlier, incorporates certain distinctions because it operates based on temporal rules.
 As we show below, the above organization of learning leads to a larger sequential memory capacity for RNNs. See Appendix \ref{app:rnn} for a discussion of the traditional RNN architecture and the learning rule, i.e., BPTT. In the following, we refer to traditional RNNs as `n\"aive RNN'.

\subsection{Performance Measure}
In this paper, we adopt an online evaluation approach, training our model on an input window of past $T$ training samples and assessing its effectiveness in predicting the $(T+1)^{th}$ sample, as we slide over the time series data. To compute the prediction accuracy in an online manner, we maintain a sliding window of $win$ samples:
\begin{equation}
    \text{Prediction Error} = 1 - \frac{1}{win} \sum_{t=t_1}^{t_1+win} \mathds{I} (\hat{x}_t = x_t)
\end{equation}
In our subsequent experiments, we choose a sliding window, $win = 1000$, as it resulted in a less noisy estimate of the performance.
\section{Simulation Setup}
\label{sec:simulation}
        
        

\begin{figure*}[!t]
    \centering
    \includegraphics[width=0.75\textwidth]{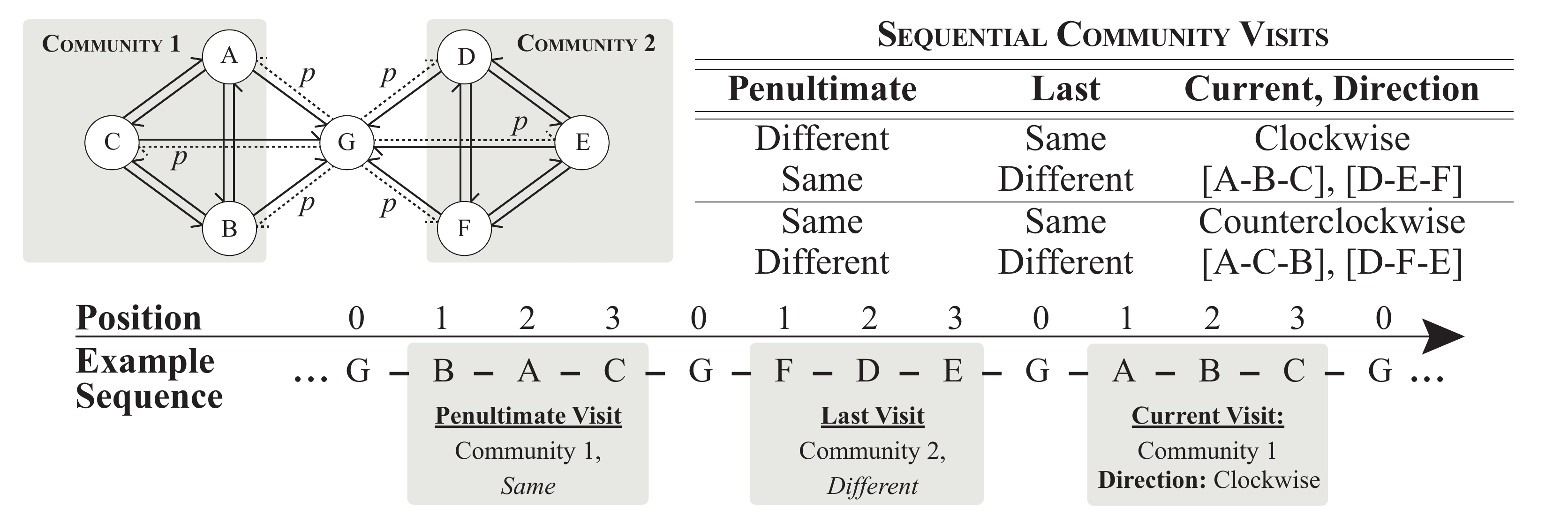}
    \caption{\textbf{Token sequence generation in the synthetic environment.} 
\textit{Top Left:} Transition graph showing two communities, \{A, B, C\} and \{D, E, F\}, connected via a central hub token G. Solid arrows represent deterministic intra-community transitions and returns to G; dashed arrows indicate uniform random transitions ($p = \frac{1}{6}$) from G to any community entry token. 
\textit{Top Right:} Rule table for determining traversal direction (clockwise or counterclockwise) within a community, based on comparisons between the current, most recent, and penultimate community visits. Each is labeled \textit{Same} or \textit{Different} with respect to the current community. 
\textit{Bottom:} Sample token sequence. Each token is annotated with its position in the current traversal cycle: Position 0 for hub token G, and Positions 1–3 for ordered tokens within a community. See Appendix Figure~\ref{fig:sim_ex} for additional traversal examples.}

    \label{fig:simulation}
\end{figure*}

To study how recurrent models behave under memory constraints, we construct a synthetic sequence modeling environment where we can precisely control the temporal dependencies required for successful prediction. This allows us to isolate and probe the limitations of models with restricted context windows in a structured and interpretable way. Specifically, we sought to develop an environment that: (a) contains temporal regularities on multiple timescales ; (b) includes a hierarchical structure that allows for an easy expansion of the model in future iterations; and (c) can be readily translated to an experimental task that may be used in human studies to test temporal learning skills without relying on previous knowledge. To fit all three conditions, we created an environment with a “community structure”, as described below. 

Figure~\ref{fig:simulation} illustrates the underlying transition graph for our simulation. The system comprises two distinct \emph{communities} of tokens: Community~1 with tokens \{A, B, C\} and Community~2 with tokens \{D, E, F\}. These communities are cyclically connected and can be traversed either in a clockwise (ABC, DEF) or counterclockwise (ACB, DFE) direction. Another token, G, acts as a hub or bridge connecting the two communities. Transitions always return to G after one full traversal of a community.

At each transition from node G, the system randomly selects one of the six tokens \{A, B, C, D, E, F\} with equal probability $p = \frac{1}{6}$. The selected token determines both which community to enter (Community~1 for A, B, C; Community~2 for D, E, F) and where the traversal will begin within that community. Once inside a community, the direction of traversal—clockwise or counterclockwise—is determined by the identity of the two most recent community visits relative to the current one. If both the last and penultimate visits were to the same community as the current visit, the system follows a counterclockwise path. Otherwise, it proceeds clockwise. This design ensures that the traversal pattern is not fully observable from local context alone and requires memory of recent transitions to resolve correctly. Moreover, it also highlights the hierarchical nature of the task, since the critical factor determining the direction of transitions within a community is previous transitions between communities (rather than local transitions between states).

To determine the traversal direction within a community, the system must track \textit{the last seven tokens}: one for the current token, four to identify the previous community, and two to identify the penultimate community. For instance, consider the example shown in Figure \ref{fig:simulation}. During the current visit community $2$ is entered via Token A, and the system needs to decide whether to move clockwise (next token: B) or counterclockwise (next token: C). The last four tokens—G, F, D, E—indicate that the most recent visit was within Community 2. By recalling the $6^{th}$ and $7^{th}$ tokens, such as G and B, the system can infer that the penultimate visit occurred within the same community, Community 1. Therefore, the system follows the clockwise transition, selecting B as the next token.

While the complexity of the task can be easily scaled by increasing the number of past visits required to determine transitions, we restrict our experiments to this seven-token memory setup. This setting provides a well-defined lower bound on the context window required for accurate prediction, enabling a direct evaluation of whether a model can learn the implicit transition rules with limited resources.

Note that the environment is partially stochastic: although three out of every four tokens in a community cycle are deterministically predictable, the token following G (i.e., the community entry point) is selected randomly. As a result, even an ideal learner can at most only achieve an optimal accuracy of $\frac{3 + \frac{1}{6}}{4} = 79.17\%$.


\subsection{Rationale for choosing the above simulation}

As noted earlier,  our simulation task was designed to allow comparisons to human temporal learning capabilities while performing a corresponding experimental task. To mirror the core structure of this simulation setup, we employed SRTT—a behavioral paradigm widely used in cognitive neuroscience to investigate implicit sequence learning (described in the next section). This choice enables a direct comparison between machine learning models and human learning under similar constraints, while offering precise control over structural dependencies in the data. We believe this setup is particularly suitable for the following reasons:
\begin{itemize}
    \item \textbf{Alignment with Human Experiments:} Similar to our simulation goals, the SRTT is known to tap into temporal pattern learning—as long as participants are not explicitly told that a pattern exists \citep{lerner2019sleep}. Through the SRTT paradigm, human participants undergo a controlled training period where their implicit learning of the underlying pattern is assessed before their explicit awareness of the pattern is tested. As such, the SRTT offers an aligned method of training and testing to compare our modeling work to. 
    \item \textbf{Minimal Prior Knowledge Requirements:} Specifically, the SRTT is known to allow differentiation between visuomotor learning, reflecting the ability to quickly react to specific locations highlighted on screen, and sequential pattern detection, which reflects the ability to identify regularities in the order of appearance of those highlighted locations \citep{robertson2007serial}. Whereas the first may depend on participants’ initial visuomotor mapping skills, the second does not rely, and cannot be significantly aided, by any previous knowledge, as the pattern is fully determined by the task parameters (unlike, for example, pattern detection tasks involving word associations that inherently depend on existing language skills). Indeed, it is known that the hippocampus, one of the main brain regions involved in encoding \textit{new} information, is inherently involved in the pattern detection aspect of the task but less so in the visuomotor aspect \citep{albouy2013interaction, curran1997higher, destrebecqz2005neural}. Thus, while it is always true that humans bring some prior knowledge to any situation, the specific knowledge participants bring to this task (e.g., visual recognition of shapes;  how to interface with a computer and keyboard) does not correspond in any way to the critical hidden pattern embedded in this task.
\end{itemize}


\section{Machines and Humans after Limited Training--Without Offline Period}

\subsection{Concepts Learned by Humans}
\label{sec:human_study}

\begin{figure*}[!ht]
  \begin{center}
    \begin{subfigure}[t]{\textwidth}
      \centering
      \includegraphics[width=.7\textwidth]{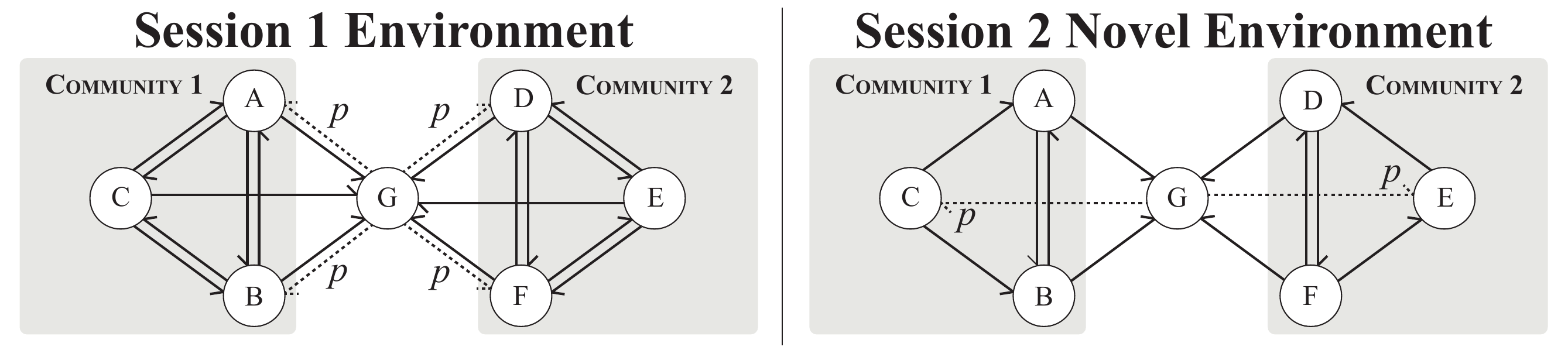}
    \end{subfigure}\\
    \quad
    \begin{subfigure}[t]{\textwidth}
      \centering
      \includegraphics[width=.8\textwidth]{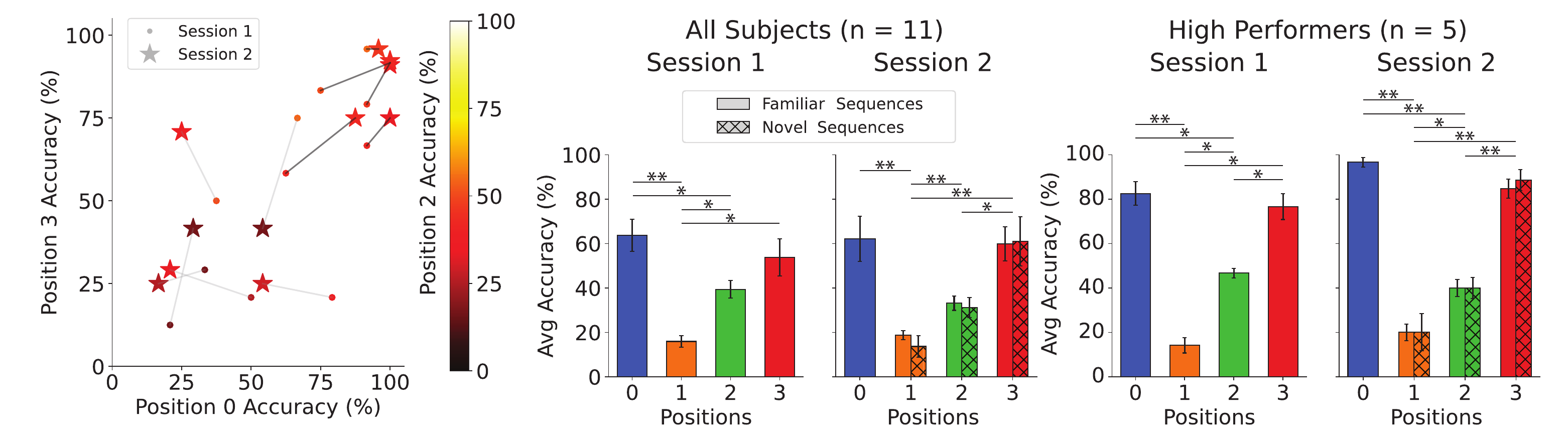}
    \end{subfigure}
  \end{center}
  \caption{\textbf{Humans learn the community structure and successfully utilize the learned structure in a novel environment.}
  \textit{Top:} Graphs showing connectivity for the Session 1 environment (state transition probability $p=\frac{1}{4}$) and the Session 2 novel environment ($p=\frac{1}{2}$). \textit{Bottom Column 1}: Individual subject performance for sessions 1 and 2 shown as a function of average accuracies for predictions made at positions 0, 2, and 3. High performing subjects cluster in the top right quadrant (n = 5, dark line connecting same-subject sessions). Average accuracy rates for all subjects tested using session 1 environment (\textit{Bottom Column 2}) and and tested using full state transitions ($p=\frac{1}{6}$) in session 2 (\textit{Bottom Column 3}). Average accuracy rates for high performing subjects (n = 5) during session 1 (\textit{Bottom Column 4}) and session 2 (\textit{Bottom Column 5}). * : p $<$ 0.05; ** : p $<$ 0.01.
    }
  \label{fig:PilotHumanStudy_GT}
\end{figure*}
In this section, we conducted a pilot study with human subjects (n = 11) to investigate learning efficiency, performance, and generalization of a temporal task equivalent to the one trained by the computational models. We aimed to establish human baselines of time-series learning and understand what cognitive abstractions occur during wake. This pilot study was developed, administered, and analyzed from November 2024 to March 2025 and was approved by the Institutional Review Board (IRB) to ensure ethical compliance. To train subjects, we used the SRTT, an established method to investigate implicit learning of underlying temporal patterns \citep{Fischer2006}. 

In short, in each trial participants observe a star shortly appearing in one of 7 possible boxes and need to indicate the star location by pressing one of the 7 corresponding buttons. After a response is given, the star reappears in the next location. The time-series of locations follows the same rules as in the computerized environment learned by the model, though participants are never explicitly told a pattern exists. Right after training, participants are tested for their knowledge of the temporal pattern through a generation task, where they need to actively predict the location of the next star after observing a series of transitions. They are tested again in a second session 10 minutes later (see Appendix Figure \ref{fig:psychopy} for detailed methods of cognitive task). The first session uses a state transition probability of $1/4$  as shown in the session 1 environment (Figure \ref{fig:PilotHumanStudy_GT}). The second session differs by allowing all state transitions ($p = 1/6$), where the novel environment sequences are presented to subjects alongside the familiar session 1 sequences (Figure \ref{fig:simulation}).

Following the 13 training blocks of SRTT (201 trials per block) in session 1, results show implicit learning evidenced in the gradually decreasing average reaction time across normal SRTT blocks and in the increased average reaction time in catch trials compared to non-modified trials (see Appendix Figure \ref{fig:srttresults}). As shown in Figure \ref{fig:PilotHumanStudy_GT}, Generation Task performance varied based on the sequence position (96 total trials per session; 24 trials per position), where some subjects show more explicit learning than others. High performing subjects (n = 5; criteria: average total accuracy $\geq$ 50\%) robustly detect the boundaries of individual communities and are able to deduce the final token within a community visit in both sessions, as evidenced by the high accuracy rates for responses predicting positions 0 and 3. This high-performing group does not achieve the same high accuracy rates when making a prediction at position 2, suggesting they were unaware of the temporal rule determining direction within a community visit. Finally, subjects do not show different performances in session 2 for familiar sequences compared to novel sequences when separated in analysis. This suggests that subjects successfully generalized the pattern rules learned during the limited training session; future work is required, though, to better understand generalization strategies in humans.
In the pilot study we ran on humans using the SRTT with our synthetic environment, we found that subjects—particularly high performers—in this pilot study are capable of predicting tokens at positions 0 and 3 with high accuracy but struggle to surpass the chance accuracy at position 2. Interestingly, humans show an immediate ability to generalize during wakeful learning and are, thus, a model system to inspire novel learning mechanisms for artificial networks. Though only a preliminary result, humans in our pilot study appear to employ a cognitive chunking mechanism during wakeful learning to capture short-range temporal regularities recognizing community boundaries without the aid of prior knowledge. Subjects, however, apparently fail to capture the long-range dependencies required to achieve optimal accuracy, suggesting the formation and application of cognitive chunks has complex limitations. We hypothesize that allowing subjects to sleep between sessions 1 and 2 will facilitate sleep-dependent memory consolidation as described by the temporal scaffolding hypothesis and, thus, show a larger proportion of subjects meeting `high performer' criteria and, importantly, \emph{an ability to learn the direction rule (depicted by position 2 performance)}. 


\subsection{Concepts Learned by N{\"a}ive RNNs}
\label{sec:taditional_rnn}

\begin{figure}[!ht]
    \centering
    \includegraphics[width=.52\textwidth]{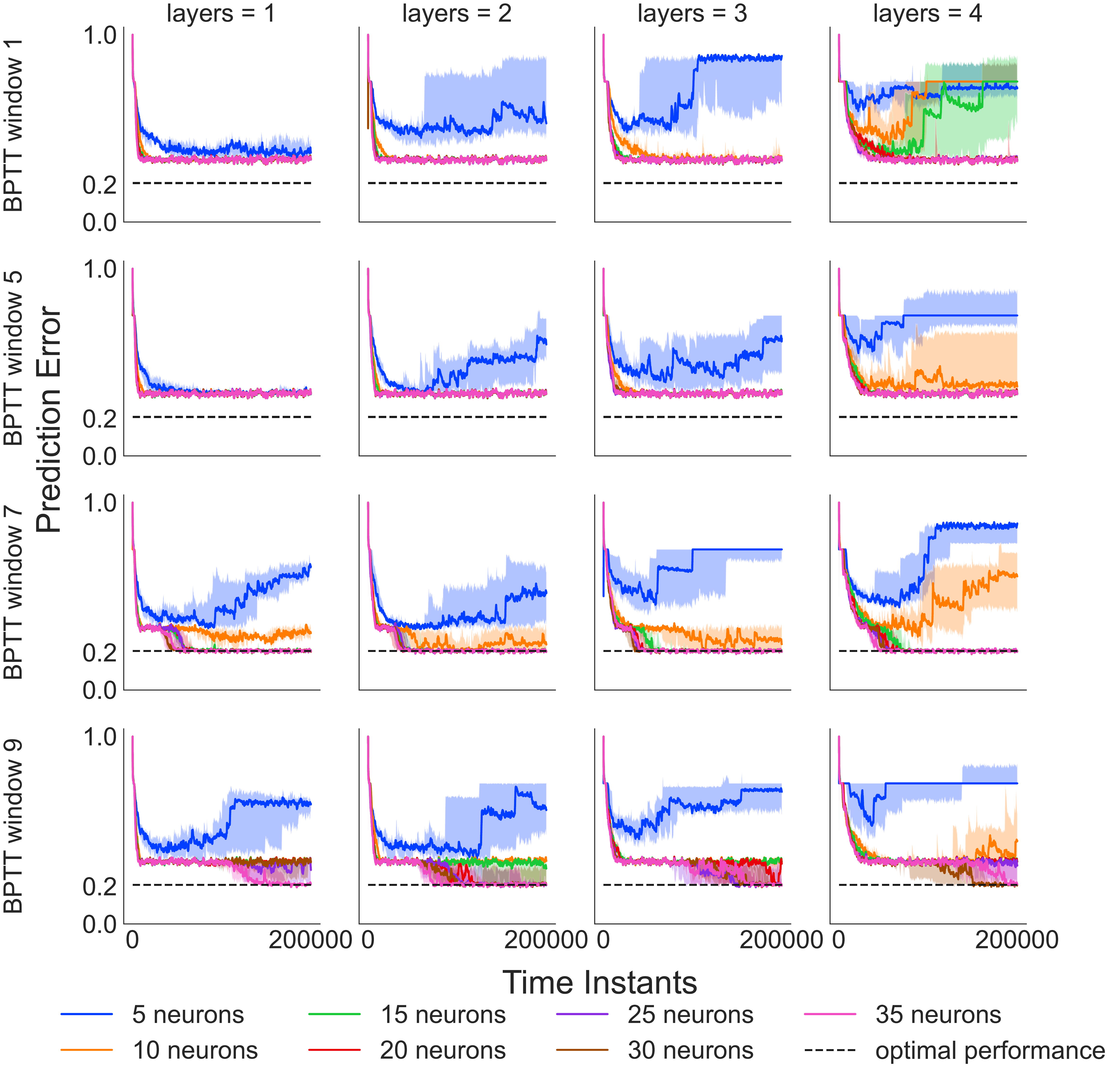}
    \caption{\textbf{Performance of RNN on our simulation setup.} A n{\"a}ive RNN is trained on the simulation setup described in Section~\ref{sec:simulation} with varying neurons, layers and BPTT window. Studies show that an RNN needs at least $15$ neurons, a single layer, and a $7$-step BPTT input window to achieve the optimal performance. The error bars (interquartile range) are shown as shaded region on both sides of each solid curve (median prediction error).
    }
    \label{fig:minimal_rnn}
\end{figure}

In this section, we train a n{\"a}ive RNN on the simulated time series data set described in Section \ref{sec:simulation} with ablation studies by varying the number of layers, the number of nodes in each layer, and the length of the BPTT window. Each token is one-hot encoded before using as an input to the RNN. After each BPTT step the hidden state was initialized to the previous hidden state and the experiments were repeated $10$ times to obtain the error bars. As shown in Figure \ref{fig:minimal_rnn}, an RNN requires at least a BPTT window of length $7$ to achieve optimal performance. Note that our designed time-series system requires a memory of the last $7$ tokens to perfectly determine the next token in the communities. This experiment shows that \textbf{ a traditional RNN does not remember tokens outside its input window.} 

\begin{figure}
  \centering
  \includegraphics[width=0.4\textwidth]{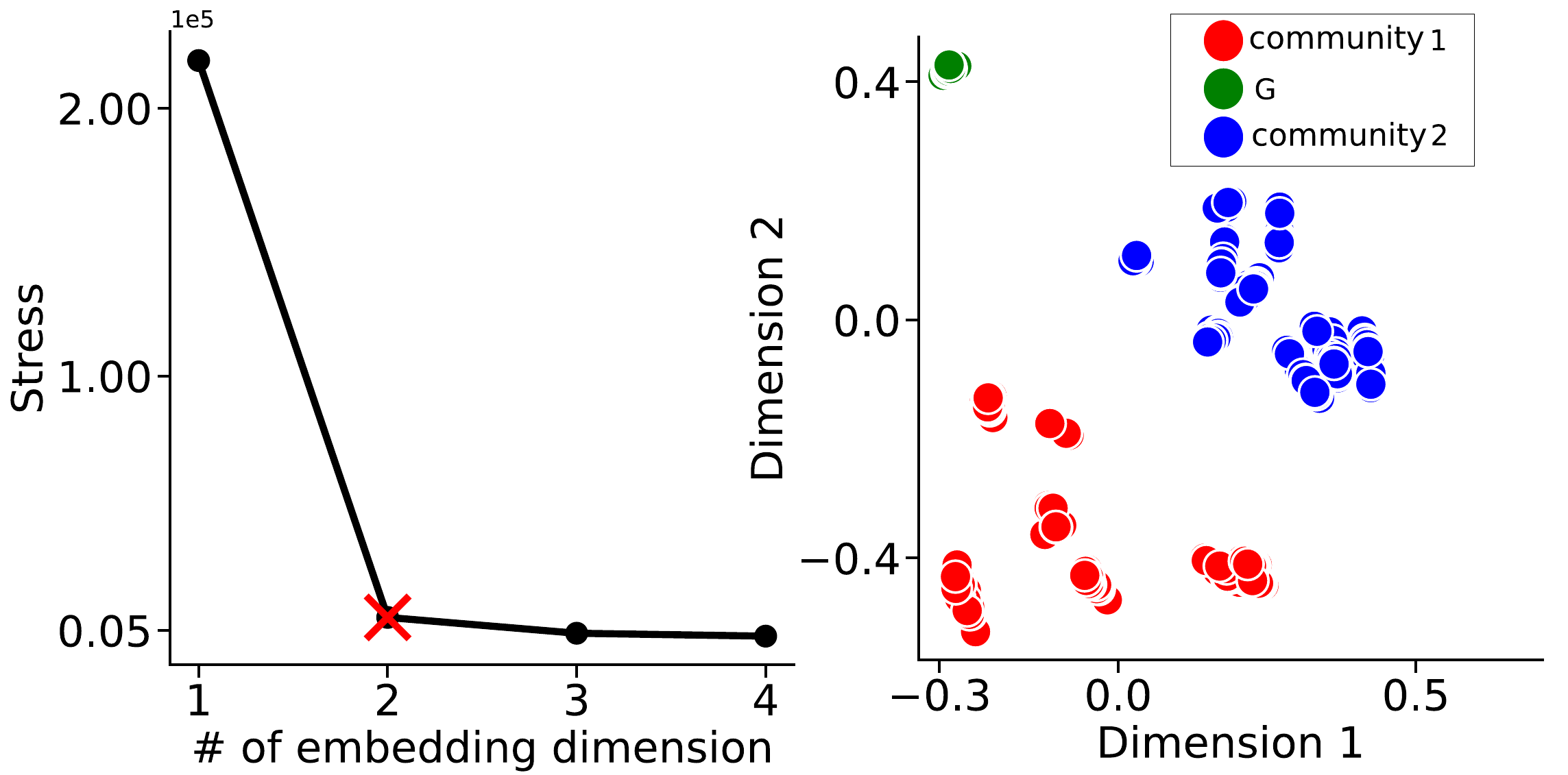}
  \caption{\textbf{A naïve RNN learns the community structure.} 
  \textit{Left:} Scree plot showing the optimal embedding dimension (knee) at $2$ (red cross). 
  \textit{Right:} Visualization of the low-dimensional manifold of RNN hidden states, obtained via Multidimensional Scaling (MDS) using cosine distance. }
  \label{fig:naive_rnn}
\end{figure}



 Next, we experiment with the hidden states of an RNN layer operating with suboptimal resources ($10$ nodes and $1$ BPPT input window, for example) and observe the concepts that a n{\"a}ive RNN learns. As shown in Figure \ref{fig:naive_rnn}, the hidden states of a n{\"a}ive RNN after a brief training session are clustered according to their respective community. The hidden states are derived for each input token and their pairwise distance matrix is embedded into a two-dimensional space. This suggests that we can learn a linear classifier to detect the community and use the community information as a context tag to chunk the learning in the next training session. Figure \ref{fig:naive_rnn} bottom row column $3$ further shows that the model detects tokens at the boundary positions of a community (positions $0$ and $3$) with high accuracy. Position $0$ denotes the token just before entering a community (`G') and other positions in the community are denoted sequentially as position $1,2,3$.



\section{Our Proposed Approach}
\begin{figure*}
\centering
\begin{subfigure}{0.65\textwidth}
    \includegraphics[width=\linewidth]{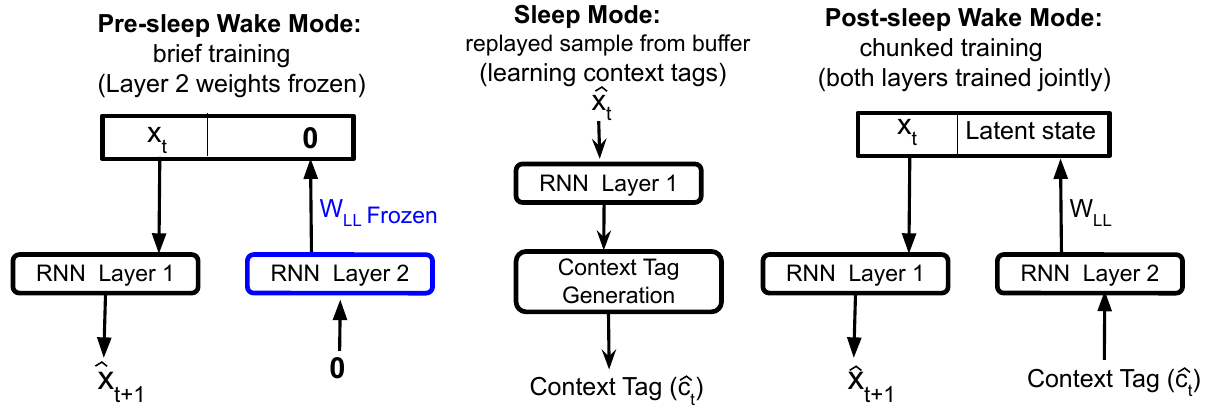}
    \caption{Block diagram showing a training procedure of a two-layer chunked RNN.}
    \label{fig:subfig1}
\end{subfigure}
\hfill
\begin{subfigure}{0.3\textwidth}
    \includegraphics[width=\linewidth]{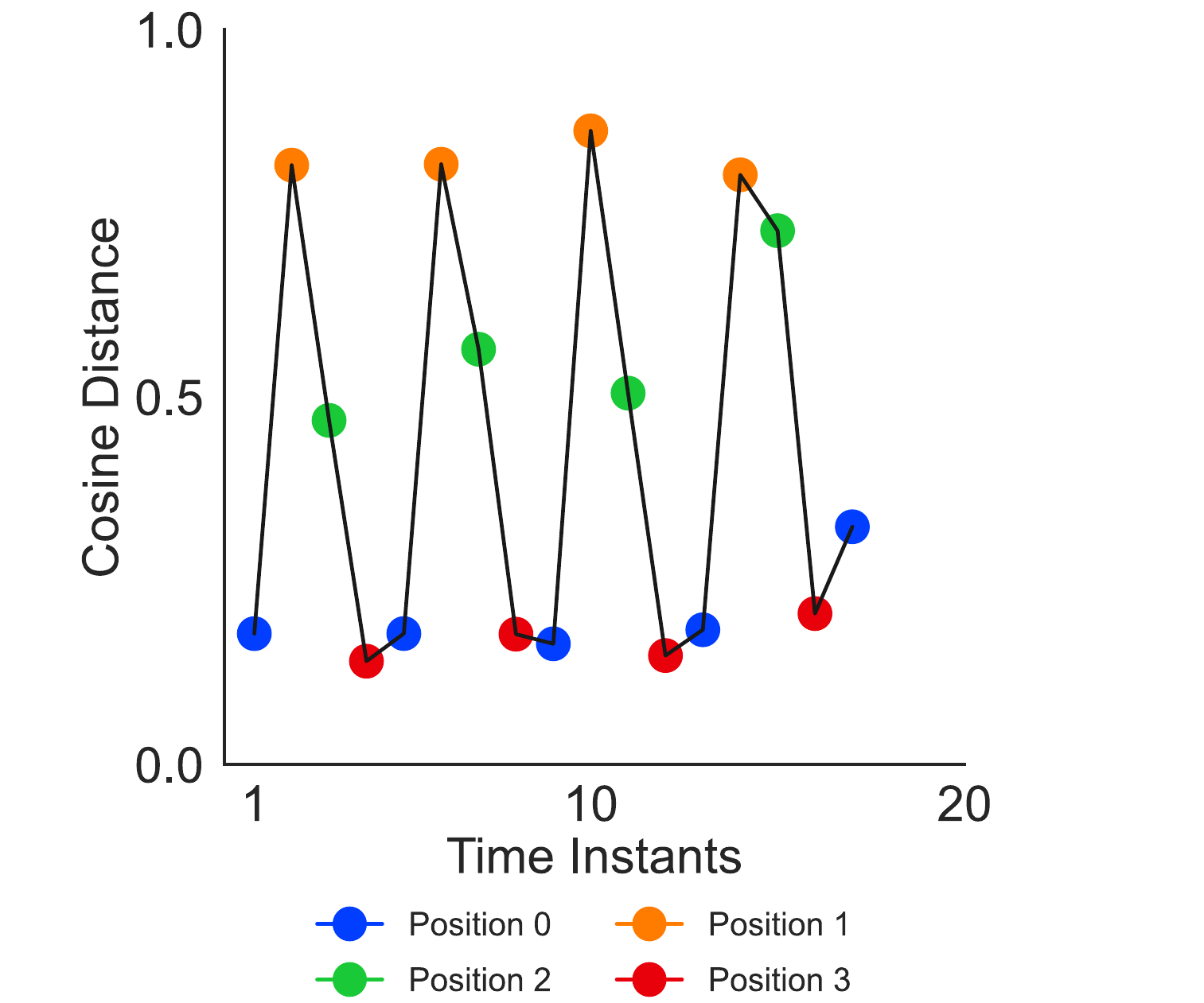}
    \caption{Context tag generated with cosine distance between the hidden states of subsequent tokens.}
    \label{fig:subfig2}
\end{subfigure}
\caption{\textbf{Proposed three-step chunked training.} (a) In this model a learner first collects experience from the environment in a pre-sleep mode, derives the context tags during sleep and finally, and uses the context tags to chunk the learning in the post-sleep mode. (b) Context tags for this study are generated using cosine distances to identify an entry into a community. }
\label{fig:block_diag}
\end{figure*}

As described in Section \ref{sec:problem}, our approach organizes the concepts learned by an RNN as context tag and chunk pairs $(c_i,\mathcal{P}_i)$. However, the characteristics that define a context tag have not been previously investigated. In this pilot study, we show a way to derive the context tags. As described in the previous section, a naive RNN learns the community structure after a brief training session. We utilize the structure learned above to derive the context tags. As shown in Figure \ref{fig:block_diag} right panel, the cosine distances between the hidden states for subsequent tokens are periodic, and the peak distance occurs at the first position of a community after 'G'. This periodicity suggests that the RNN must allocate most of its resources to encoding the peaks, while dedicating fewer resources to the tokens that follow, as they are semantically close to the peak position in the representation space. This phenomenon is also evident in Figure \ref{fig:naive_rnn} bottom row column $3$ and $4$ where the naive RNN has the lowest accuracy in predicting Position $1$. We detect the peaks of the distances and derive a mask for a training sequence of length $L$, $\mathbf{m} = \{0,1\}^L$ where the mask is $1$ whenever the beginning of a community is detected. We use the above training sequence and mask to train a small RNN ($1$ layer, $5$ hidden neurons), $g: \mathcal{S} \rightarrow \{0,1\}$, to identify whether a given token marks the beginning of a community ($1$) or not ($0$). We train this separate RNN so that when the weights in layer $1$ are fine-tuned again in step $2$, it does not affect the detection of the context tags. We use the detected Position $1$ token as a context tag in our next training step. For example, if the sequence is `ABC G DEF G CAB G', the mask will be `100 0 100 0 100 0' and the context tags are `AAAA DDDD CCCC'. In summary, a new context can be identified when there is a big change in the environmental input, followed by smaller changes. Figure \ref{fig:block_diag} shows a block diagram of our three-step training procedure. In the pre-sleep wake mode, Layer $2$ of the RNN is kept frozen while Layer $1$ goes through a brief training session with the inputs from the environment. During the sleep mode, experience gathered from the environment in a buffer are replayed to Layer $1$ and the context tags are learned in the procedure described above. In the post-sleep wake mode, Layer $2$ of the RNN is kept frozen while Layer $1$ goes through a brief training session. In the sleep mode, context tags are derived based on the learning in Layer $1$ in the previous step. In the post-sleep mode, both layers are trained jointly using the context tag and the streaming training sequence. Note that the context tag can have broader meaning in other learning setups such as multimodal learning. For example, a familiar smell or scene can often invoke a chunk of past memories in humans \citep{schacter2008searching}. We can hierarchically construct larger chunked concepts by deepening the RNN, where each $(l+1)$ th layer chunks the context tags $\mathcal{C}_l$ learned by the previous $l$ th layer. However, in this paper, we restrict our experiments to two layers of an RNN. In particular, since RNNs can be used as generative models, context tag learning can occur during a brief resting or sleep cycle without any external input from the environment, resembling the process of human sleep. However, we will further investigate the above sleep-based memory consolidation in future studies involving humans and machines. See Appendix \ref{app:pseudo-code} for a pseudo-code of our approach.


\subsection{Learning Sequential Pattern with Chunked RNNs}

\begin{figure*}[t]
    \centering
    \includegraphics[width=0.7\textwidth]{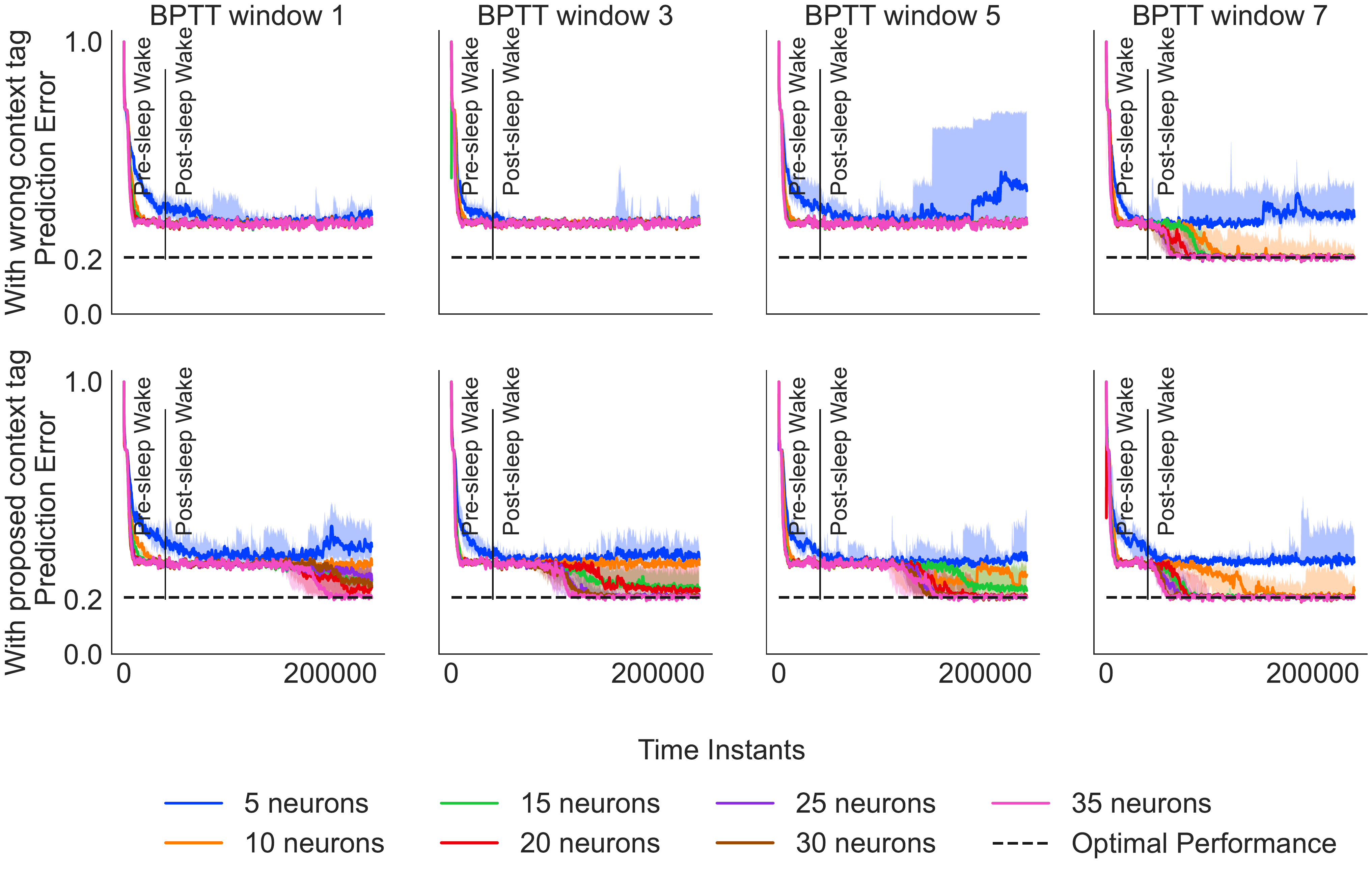}
    \caption{\textbf{Chunking enables a two-layer RNN to achieve optimal performance with fewer resources.} \textit{Top:} Ablation study where the context tag is fixed to a constant token. Without meaningful context tags, the proposed RNN architecture fails to reach optimal performance when using fewer than 7 BPTT steps. \textit{Bottom:} When the context tag indicates the boundary of a community, the proposed RNN achieves optimal performance even with one BPTT step.
    }
    \label{fig:chunked_model}
\end{figure*}

In Section \ref{sec:simulation}, we described our simulation setup in which we enter a community via any of the three tokens in the community. In this section, we train the proposed chunked RNN architecture on the above simulation setup. In Figure~\ref{fig:chunked_model} (top), we conduct an ablation study where the context tag is fixed to a constant token. In the bottom row, we train using the proposed context tag derived from community boundaries.

As shown in Figure~\ref{fig:chunked_model} (bottom), the chunked RNN achieves optimal performance with a BPTT window as small as $1$, while the na\"ive RNN requires a window of at least $7$ to achieve similar performance (see Figure~\ref{fig:minimal_rnn}). The fact that the proposed context tag enables the model to perform optimally with a significantly reduced window size highlights the effectiveness of the temporal chunking mechanism (see Figure~\ref{fig:chunked_model} top).

\subsection{Transfer Learning with Chunked RNNs}



\begin{figure*}[htbp]
    \centering
    \includegraphics[width=0.75\textwidth]{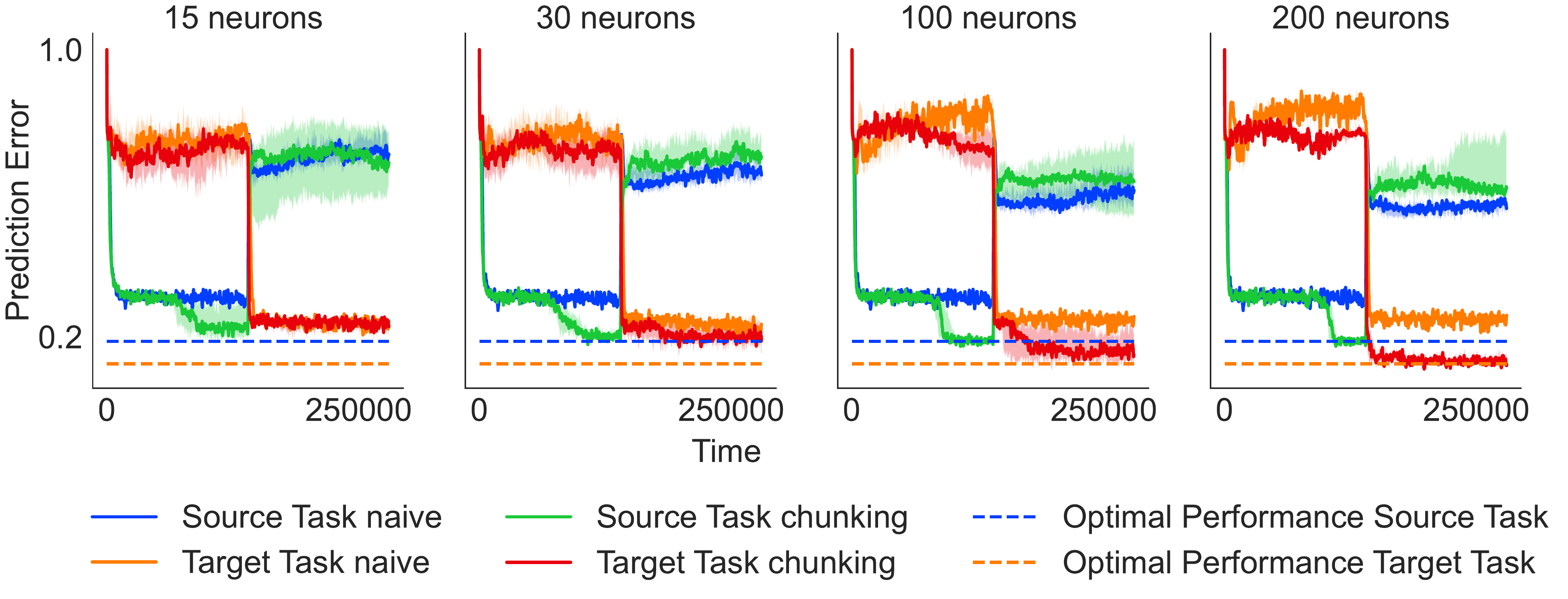}
    \caption{\textbf{RNN with chunking results in better forward transfer than a naïve RNN.} Performance on target task for chunked RNNs (red curve) improves over that of naïve RNNS (yellow curve) as the target task is introduced. 
    }
    \label{fig:chunked_model_cl}
\end{figure*}

In this section, we evaluate chunked RNNs in a transfer learning setup. We use the Session 1 environment as the source task for training and fine-tune on the Session 2 novel environment as the target task, as illustrated in the top row of Figure~\ref{fig:PilotHumanStudy_GT}. We first train a two-layer RNN on the source task in three steps as described in Figure \ref{fig:block_diag}. We use the same number of layers and nodes for the n{\"a}ive and the chunked RNN. When transitioning to the target task, Layer $2$ receives context tags learned from the source task. Both layers are trained jointly, as shown in the post-sleep wake mode in Figure \ref{fig:block_diag}. Note that layer $2$ is frozen only once for the source task and never for the target task. After a brief training session, context tags are learned for the target task from the hidden states of Layer $1$ and thereby both layers are trained jointly again. This structure enables the chunked RNN to achieve optimal performance on the target task faster than a n\"aive RNN. In particular, chunked RNNs exhibit faster learning as the number of neurons per layer increases. \emph{Chunking improves knowledge transfer from the source to the target task.}
\section{Related Work}
Improving the ability of models to learn long-range and varying temporal structures has been an active research development in both RNNs and transformers. An early idea was that hierarchy in RNN layers leads to abstraction across longer timescales \citep{hihi1995hierarchical}. However, as demonstrated in this work with a naive RNN, this does not empirically perform well when the BPTT window is too short to leverage these hierarchical relationships. Another approach,clockwork RNN, which implements a single RNN as a network of modules operating at different timescales \citep{koutnik2014clockwork}, allowing for greater flexibility in representation and neuron dynamics, but still restricted to predefined timescales. Building on this, Hierarchical Multiscale RNNs \citep{chung2016hierarchical} introduced an adaptive boundary detector to learn temporal patterns at varying scales. Our approach differs in the sense that it facilitates bidirectional passing of information between layers to learn different temporal patterns, especially when the BPTT window is too short to capture the full timescale of interest. 

In contrast to modifying the architecture, researchers have also investigated modifications to the BPTT algorithm, such as allowing the model to skip over timesteps, effectively increasing the duration for which BPTT can be applied \citep{campos2017skip, chang2017dilated}, or by constraining the weight matrices through unitary \citep{arjovsky2016unitary} or antisymmetric \citep{chang2019antisymmetricrnn} weights to mitigate unstable gradient updates and facilitate long-term modeling. These methods do not investigate learning long-term dependencies that exceed the BPTT window length. Similar efforts have been made to extend the context window for transformers by proposing strategies to reuse previous information \citep{dai2019transformer, rae2019compressive}. Alternative approaches include novel architectures, such as augmenting RNNs with external memory \citep{graves2014neural}, state-space models for longer sequence modeling \citep{gu2021efficiently}, combining the benefits of transformers and RNNs \citep{yang2023gated, peng2023rwkv}, and using inductive biases for higher-level cognition \citep{goyal2022inductive}.
\section{Discussion}
In this paper, we have studied how temporal chunking enhances the sequential pattern recognition capability in time-series models. Our proof-of-concept experiments demonstrate that chunking improves learning speed, temporal memory, and the potential for transfer learning between related tasks. 

Furthermore, our pilot study on human subjects suggests that after initial training, they can recognize community structures in a manner similar to the n{\"a}ive RNNs. However, this human study is currently limited by the small sample size and the lack of sleep condition. In the future, our aim is to conduct a complete human sleep study to investigate differences in temporal pattern learning in wake-only compared to post-sleep testing. 

Another potential limitation of our work is the assumption of some ideal conditions, including the availability of a memory buffer storing environmental inputs during sleep and the use of an additional RNN to preserve context tags from being overwritten. In the future, we will extend our work beyond these assumptions by using Layer $1$ of the chunked RNN to generate inputs from wake experiences and applying unsupervised clustering approaches to identify context tags. In future work, we plan to extend the chunked RNN architecture beyond two layers and evaluate its performance on benchmark datasets.

In addition, we plan to explore whether sleep-dependent memory mechanisms, such as temporal chunking, can enable both humans and machines to achieve learning of underlying sequence rules in a non-stationary environment. 
The codes for reproducing the experiments in this paper will be made publicly available.

\section{Acknowledgment}
This work is graciously supported by the NSF Award $\#2317706$. 

\bibliographystyle{unsrtnat}
\bibliography{ref}

\clearpage
\appendix
\section{Recurrent Neural Network}
\label{app:rnn}
RNNs are a type of neural network architecture which is used to detect patterns in a sequence of data. An RNN maintains a temporal memory in the form of a recurrent connection. We denote the hidden state and the input at time step t respectively as $H_t \in \mathbb{R}^{n\times h}$ and $X_t \in \mathbb{R}^{n\times d}$ where $n$ is the number of samples, $d$ is the
dimension of each sample, and $h$ is the number of hidden units. In addition, it has a weight matrix $\mathbf{W}_{xh} \in \mathbb{R}^{d\times h}$, hidden-state-to-hidden-state matrix $\mathbf{W}_{hh} \in \mathbb{R}^{h\times h}$, bias parameter $\mathbf{b}_h \in \mathbb{R}^{1\times h}$ and an activation function $\phi$ which is usually a ReLU or a tanh function. The equation for the hidden state can be written as:

\begin{equation}
    \mathbf{H}_t = \phi_h(\mathbf{X}_t \mathbf{W}_{xh} + \mathbf{H}_{t-1} \mathbf{W}_{hh} + \mathbf{b}_h)
\end{equation}

$L$ such hidden layers can be stacked on top of each other, and the final hidden state is passed through a linear layer to provide the output:

\begin{equation}
    \mathbf{O}_t = \phi_o (\mathbf{H}_t \mathbf{W}_{ho} + \mathbf{b}_o)
\end{equation}

\begin{figure*}[!ht]
    \centering
    \includegraphics[width=.5\textwidth]{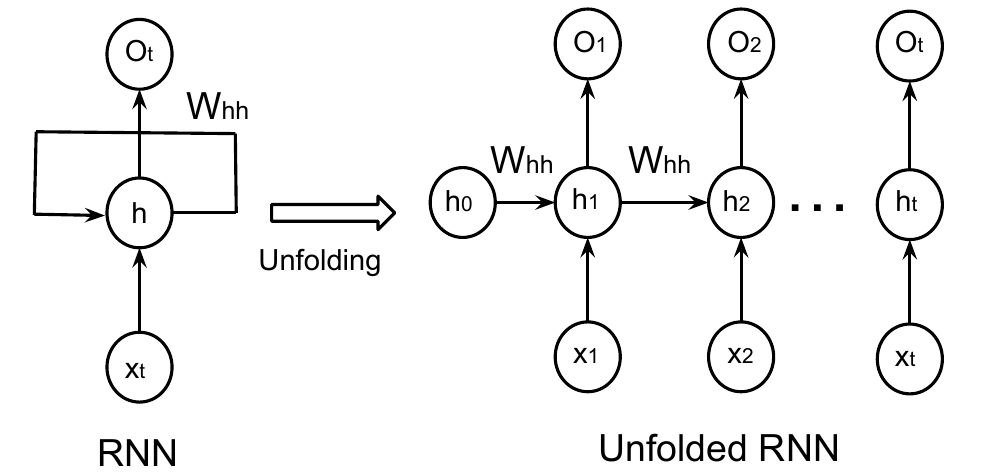}
    \caption{\textbf{Backpropagation through time.} \textit{Left:} An RNN with input $x_t$ and output $o_t$. \textit{Right:} Unfolded RNN over time steps. 
    }
    \label{fig:bptt}
\end{figure*}

\subsection{Backpropagation Through Time}
In an RNN, the recurrent weights are shared across time steps. To optimize these weights, Backpropagation Through Time (BPTT), an extension of standard backpropagation, is utilized. BPTT unfolds the network over time, allowing propagation of prediction error through the expanded computational graph (see Figure \ref{fig:bptt}). Theoretically, computing BPTT for a single time step would necessitate an infinite unfolding of the network. However, to ensure computational feasibility, a truncated version of BPTT is often employed, where the network is unfolded over a finite time window or context period. 

\section{More Simulation Examples}
\begin{figure*}[!ht]
    \centering
    \includegraphics[width=\textwidth]{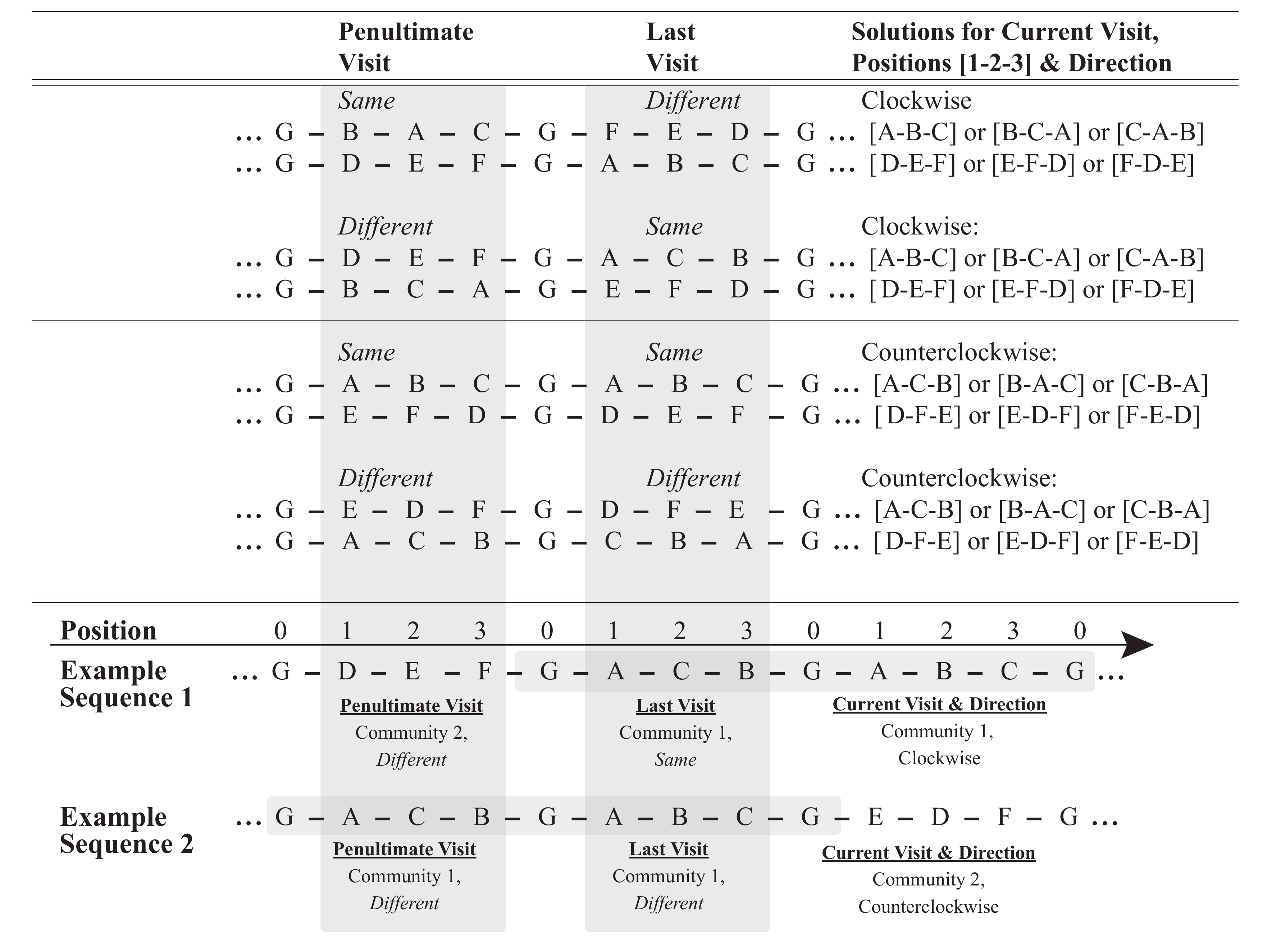}
    \caption{\textbf{Example sequence explaining our simulation setup.} The penultimate and the last visit determine the direction of traversal within the current community.
    }
    \label{fig:sim_ex}
\end{figure*}

Given the limited space in the main text, we provide here further clarification of our simulation setup. Figure \ref{fig:sim_ex} shows some additional example sequences with explicit demonstration of the community traversal rule.

\section{Pilot Human Study}
\label{app:methodshuman}
\subsection{Methods}
\begin{figure*}[!ht]
    \centering
    \includegraphics[width=\textwidth]{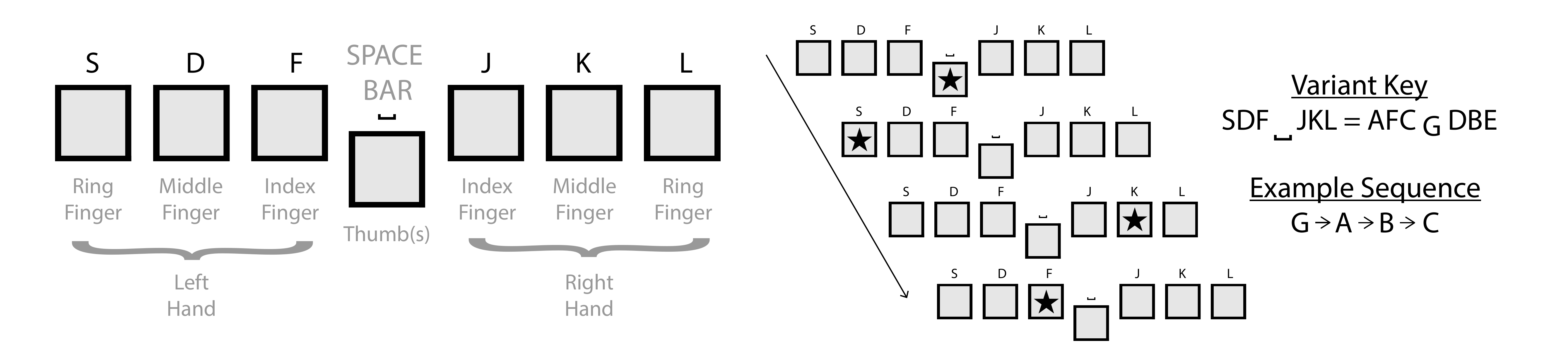}
    \caption{\textbf{Cognitive Task Setup.} \textit{Left}: Diagram of on-screen display as subjects are instructed to orient their appropriate fingers onto the designated keyboard key. \textit{Right}: Diagram showing how an example sequence (G, A, B, C) would appear as serial star presentations on a permanent background during a given task. The 'Variant Key' shows which token is designated to each keyboard key in this example. Each subject is pseudorandomly assigned to one of four variant key options, which is kept constant throughout the entire experiment for each subject.}
    \label{fig:psychopy}
\end{figure*}
\begin{figure*}[!ht]
    \centering
    \includegraphics[width=.8\textwidth]{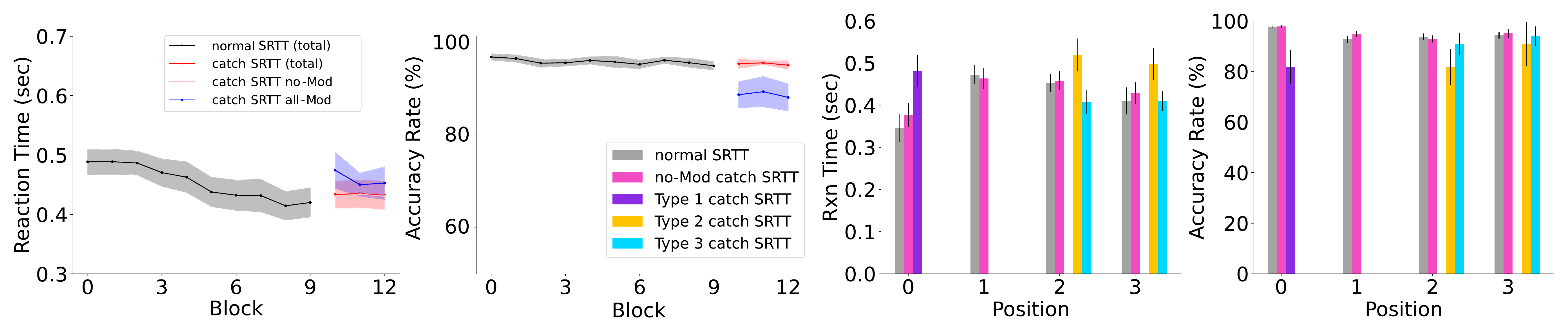}
    \caption{\textbf{SRTT Results show Implicit Learning of Community Structure during Training (Session 1).} \textit{Column 1}: Average reaction times per block of SRTT displayed in chronological order for all subjects (n = 11). \textit{Column 2}: Average response accuracies per block of SRTT for all subjects. \textit{Column 3}: Average reaction times for the final normal and catch SRTT blocks (block 9 and 12, respectively) with data separated according to responses made at positions 0, 1, 2, and 3. \textit{Column 4}: Accuracy rates for the final normal and catch SRTT blocks with data separated according to responses made at positions 0, 1, 2, and 3. Shadow boxes and error bars display $\pm$SEM values.}
    \label{fig:srttresults}
\end{figure*}
Data was collected for 11 participating subjects. Informed consent was obtained from all subjects, in accordance with an approved IRB protocol. The cognitive tasks were designed and employed on a computer screen (23.5 in. w x 13.25 in. h; 27 in. top-to-bottom corner) in the laboratory setting using PsychoPy software \citep{Peirce2019}. Throughout training and testing for all cognitive task trials, 7 white boxes were permanently presented on the screen with a black background, where each box corresponded to a correct key response on a QWERTY keyboard; colors are inverted onto a white background, as shown in Figure \ref{fig:psychopy}, for enhanced readability in this document. Responses are limited to keyboard keys S, D, F, J, K, L, and the space bar; each token corresponds to a single key. The G token is always designated as the space bar. To limit biases associated with reaction times by the dominant-hand, each token from a single community is assigned to a non-adjacent key. As seen in the Figure \ref{fig:psychopy} example sequence, community 1 tokens A, B, and C correspond to stars presented in boxes labeled as S, K, and F, respectively.

The SRTT allows researchers to gather implicit measures of learning by comparing the subject’s reaction times and accuracy rates in \textit{normal} trials, where the true sequence is maintained, compared to \textit{catch} trials, where a token in the sequence is altered to break a particular sequence rule. For a single SRTT trial, a star appears in one of the on-screen boxes for 250 msec., and subjects are instructed to respond as quickly and accurately as possible by pressing the correct corresponding key for each stimulus presentation. The serial star presentations appeared in a designated box location following a sequence governed by the temporal rules using limited state transitions (session 1) or full state transitions (session 2). If subjects respond to a star stimulus by pressing the incorrect key, a red 'X' briefly appeared on screen as error feedback before proceeding to the next trial. 

All subjects first completed a training session consisting of 10 blocks normal SRTT (201 trials per block), followed by 3 blocks catch SRTT (201 total trials per block; 9 catch trials per block); each consecutive block was separated by a 30 sec. break. Three types of catch trials were employed in each block of catch SRTT to probe which temporal rules were learned. Type 1 catch modified ‘G’ at position 0 to be one of the six alternate tokens which did not appear in the sequence immediately before (position 3) or after (position 1) the altered position 0 token. Type 2 catch modified a community token at position 2 or position 3 to be one of the three tokens from the opposite community. Type 3 catch switched the tokens at position 2 and position 3 to impose the opposite directionality within the current community visit. Each block of catch SRTT included 3 repetitions of each catch type to maintain a low catch frequency relative to non-modified trials per block to minimize the likelihood that subjects would learn modified sequences as a novel task. 

Immediately after completion of the 13 training blocks of SRTT in session 1, subjects perform the Generation Task where they are instructed to observe a sequence of star presentations in the same 7-box format as described in Figure \ref{fig:psychopy}. Subjects are instructed to observe a short sequence of star presentations (10 to 14 stars) without pressing any keys in response to the stars. Following the final star presentation for a given trial, a white ‘?’ appears on screen prompting the subject to press the key corresponding to the box where they predict the next star will appear. Subjects complete 2 blocks of Generation Task (48 trials per block) in the first session, followed by a 10 min. period where subjects are instructed to rest while remaining awake. All subjects then complete testing session 2 consisting of 2 blocks Generation Task, then 1 block normal SRTT, and 1 block of catch SRTT. Finally, subjects were explicitly asked if they detected any pattern(s) in the sequence and, if so, what pattern(s). In total, both sessions were completed in approximately 2 hours.

Response accuracies and reaction times were recorded per trial for all SRTT and Generation Task blocks. To account for any delayed reactions due to subject inattention, all SRTT data-points per subject were first filtered so that any reaction times exceeding 5 sec. were excluded. From this filtered sample, mean and standard deviation were calculated per subject and any data-points exceeding the mean + 2.5 standard deviations were excluded from the final analysis. All data-points were included in the Generation Task analysis. Statistical tests performed include repeated measures one-way ANOVA with Bonferroni post hoc testing using python package 'statsmodels'. For session 2, subject responses to familiar sequences (17 out of 24 trials per position) and novel sequences (7 out of 24 trials per position) were separated during analysis to assess generalization; responses to familiar source and novel target sequences per subject were combined in session 2 statistical analyses to maintain equal group sizes.

\subsection{Serial Reaction Time Task Results}
The average reaction times for all normal SRTT trials steadily decrease across training blocks indicating subjects familiarized themselves with the task and improved their procedural learning by the end of normal SRTT training (Figure \ref{fig:srttresults}). When all three types of catch trials are combined (blue line; 12 trials/block), subjects show a trend of reacting slightly slower and with reduced accuracy suggesting some implicit learning occurred. To better understand the responses to each catch type according to the positional group, the final blocks of normal and catch SRTT (blocks 9 and 12, respectively) are compared in the Figure  \ref{fig:srttresults} bar plots. Subjects show robust learning of the repeating G token at position 0 indicated by the increased average reaction time and decreased accuracy rate in Type 0 catch trials compared to non-modified catch trials in the same block and to the final block of normal SRTT trials. Subjects also show implicit learning of the community structure, where Type 2 catch trials at positions 2 and 3 appear to have an increased average reaction time. There is no evidence that subjects implicitly learned the temporal rule determining direction within a community visit as indicated by reaction time and accuracy rates at positions 2 and 3 for Type 3 catch trials compared to non-modified catch SRTT trials and normal SRTT trials. 


\section{pseudo-code}
\label{app:pseudo-code}
\begin{algorithm}[!hbp]
  \caption{Initial pre-sleep wake time training, Offline Chunking and post-sleep wake time context-Tagged RNN Training}
  \label{alg:chunked_rnn}
\begin{algorithmic}[1]
  \Require
  \Statex (1) Sequence $\{\mathbf{x}_1, \mathbf{x}_2, \dots, \mathbf{x}_{T_2}\}$ \Comment{Input training sequence}
  \Statex (2) BPTT window $w$ \Comment{Backpropagation window}
  \Statex (3) Base RNN model $f_{\text{base}}$ \Comment{Trained with truncated BPTT}
  
  \Ensure
  \Statex Trained model $f_{\text{chunked}}$ using context-tagged input

  \Function{TrainWithChunking}{$\mathbf{X}$, $f_{\text{base}}$}

    \Comment{\textbf{Phase 1: Wake Phase – Initial Training of Base RNN}}
    \For{$t = 1$ to $T_1$}
      \State Predict $\hat{\mathbf{x}}_{t+1} \gets f_{\text{base}}(\mathbf{x}_{t-w+1}, \dots, \mathbf{x}_t)$
      \State Compute loss $\mathcal{L}_t = \text{CrossEntropy}(\hat{\mathbf{x}}_{t+1}, \mathbf{x}_{t+1})$
      \State Update model via BPTT over window $w$
    \EndFor

    \Comment{\textbf{Phase 2: Sleep Phase – Chunk Discovery and Tagging}}
    \State Initialize memory buffer $B \gets \{\mathbf{x}_1, \mathbf{x}_2, \dots, \mathbf{x}_{T_1}\}$

    \For{$t = 1$ to $T_1$}
      \State $h_t \gets f_{\text{base}}
      (x_t)$ 
      \State $h_{t+1} \gets f_{\text{base}}
      (x_{t+1})$ \Comment{Detect peak}
      \If{\Call{CosineDistance}{$h_t$, $h_{t-1}$} $>$\Call{CosineDistance}{$h_t$, $h_{t+1}$}} 
         \State Predict $\hat{m}_{t} \gets f_{\text{context}}(\mathbf{x}_{t})$
         \Comment{Train an RNN to predict the context tags}
      \State Compute loss $\mathcal{L}_t = \text{CrossEntropy}(\hat{m}_{t}, 1)$
      \State Update model
    \Else 
        
        \State Predict $\hat{m}_{t} \gets f_{\text{context}}(\mathbf{x}_{t})$
      \State Compute loss $\mathcal{L}_t = \text{CrossEntropy}(\hat{m}_{t}, 0)$
      \State Update model
      \EndIf
    \EndFor

    \Comment{\textbf{Phase 3: Post-Sleep Training with Context Tags}}
    \State $c_0 \gets \mathbf{x}_0$ 
    \For{$t = T_1+1$ to $T_2$}
    \State $c_t \gets c_{t-1}$
    
    \State Predict $\hat{\mathbf{x}}_{t+1} \gets f_{\text{chunked}}
    ((\mathbf{x}_{t-w+1}, c_{t-w+1}), \dots, (\mathbf{x}_t, c_t))$
    \State Compute loss $\mathcal{L}_t = \text{CrossEntropy}(\hat{\mathbf{x}}_{t+1}, \mathbf{x}_{t+1})$
      \State Update model via BPTT over window $w$\\

      \State $\hat{m}_t \gets f_{\text{context}}(\mathbf{x}_{t})$

    \If{$\hat{m}(t) = 1$}
        \State $c_t \gets x_t$
    \EndIf
    \EndFor

    \State \Return $f_{\text{chunked}}$

  \EndFunction
\end{algorithmic}
\end{algorithm}



\end{document}